\documentclass{article}

\usepackage{arxiv}

\usepackage[utf8]{inputenc} 
\usepackage[T1]{fontenc}    
\usepackage{hyperref}       
\usepackage{url}            
\usepackage{booktabs}       
\usepackage{amsfonts}       
\usepackage{nicefrac}       
\usepackage{microtype}      
\usepackage{lipsum}		
\usepackage{graphicx}
\usepackage{natbib}
\usepackage{doi}

\title{RGB-D Neural Radiance Fields: Local Sampling for Faster Training}

\author{ \href{https://orcid.org/0000-0001-9046-0876}{\includegraphics[scale=0.06]{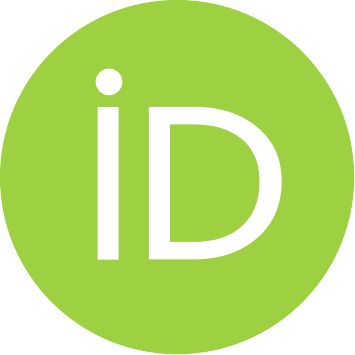}\hspace{1mm}Arnab Dey} \\
    I3S-CNRS/Universit\'e Cote d'Azur\\
    Sophia-Antipolis, France\\
	\texttt{adey@i3s.unice.fr} \\
	\And
	\href{https://orcid.org/0000-0002-3959-3195}{\includegraphics[scale=0.06]{orcid.pdf}\hspace{1mm}Andrew I. Comport} \\
	I3S-CNRS/Universit\'e Cote d'Azur\\
    Sophia-Antipolis, France\\
	\texttt{Andrew.Comport@cnrs.fr} \\
}

\renewcommand{\shorttitle}{Preprint}

\hypersetup{
pdftitle={A template for the arxiv style},
pdfsubject={q-bio.NC, q-bio.QM},
pdfauthor={David S.~Hippocampus, Elias D.~Striatum},
pdfkeywords={First keyword, Second keyword, More},
}

\begin{document}
\maketitle
\begin{figure}[ht]
  \centering
 \includegraphics[width=1\linewidth]{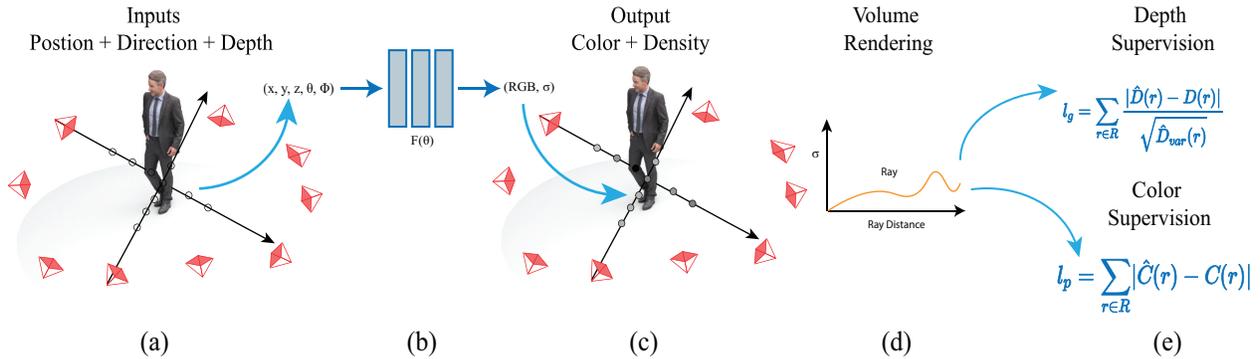}
 \centering
  \caption{The RGB-D NeRF training process. \textbf{(a)} Depth guided sampling. \textbf{(b)} The input to the network is 5D coordinate. \textbf{(c)} The network outputs volume density and color for each sample. \textbf{(d)} The color and depth of a ray generated using classic volume rendering. \textbf{(e)} The network is optimized using color and depth loss. }
\label{fig:teaser}
\end{figure}
\begin{abstract}
	Learning a 3D representation of a scene has been a challenging problem for decades in computer vision. Recent advances in implicit neural representation from images using neural radiance fields(NeRF) have shown promising results. Some of the limitations of previous NeRF based methods include longer training time, and inaccurate underlying geometry. The proposed method takes advantage of RGB-D data to reduce training time by leveraging depth sensing to improve local sampling. This paper proposes a depth-guided local sampling strategy and a smaller neural network architecture to achieve faster training time without compromising quality.
\end{abstract}


\section{Introduction and related work}
Learning the 3D representation(shape and texture) of a scene is important for novel view synthesis, 3D human modeling, virtual reality, etc. Recent advancement in neural scene representations, more specifically NeRF\cite{mildenhall2020nerf}, showed that neural networks can be used to encode high-quality images of 3D scenes. NeRF-based methods use two multilayer perceptrons(coarse and fine) to learn radiance and volume density from RGB images and their corresponding camera poses.\par
Although NeRF-like methods can produce high-quality novel views of scenes, they are too expensive to train. NeRF uses a classic volume rendering technique to compute the color of the pixels by placing 256 samples along each viewing ray. Each of those samples need a full network evaluation to compute a color. Recently, \cite{neff2021donerf} proposed real-time rendering by limiting the number of samples. They used an oracle network and ground-truth depth to predict relevant sampling locations on rays. Their method is limited to forward-facing scenes and poses belonging to a view cell. Alternatively, \cite{deng2021depth} uses sparse depth supervision generated by a Structure-from-motion(SfM) algorithm to optimize the network using color and depth information together, allowing them to use fewer input views. \cite{sucar2021imap} achieved real-time SLAM based on NeRF by using a smaller network, lower resolution inputs and removed the viewing direction. 
The method proposed here uses local sampling based on a depth sensor to reduce the number of samples and replaces the coarse NeRF network. This study aims to prove that faster training time can be achieved by local sampling without limiting scene representation quality while using a single network.

\section{Local sampling}
NeRF-like methods estimate pixel color using classic volume rendering, alpha composition \cite{mildenhall2020nerf} of sample volume density, and color to generate the final rendering. Samples with a higher volume density have a greater contribution to the final color. The proposed local sampling places fewer samples only on the relevant part of the rays given depth information.
\begin{figure}[ht]
  \centering
  \includegraphics[width=.8\linewidth]{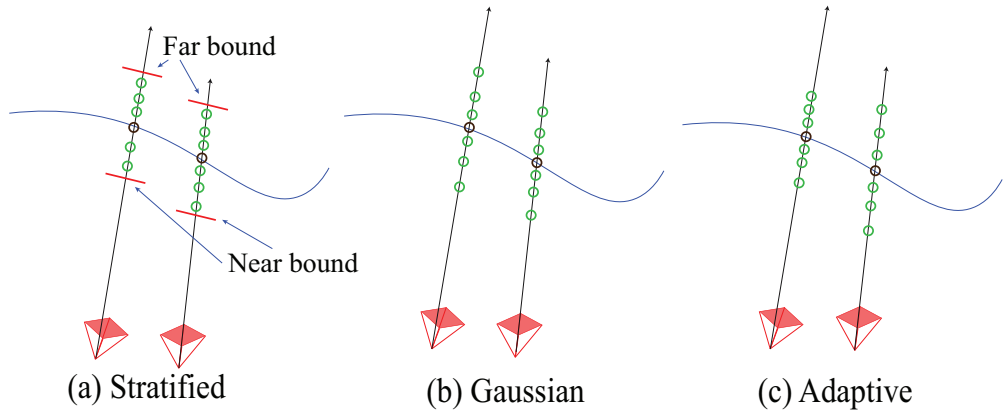}
  \caption{\label{fig:firstExample}Visualization of three different sampling strategies. Black lines represent rays coming from camera and circles are samples.}
\end{figure}
\begin{figure}[ht]
  \centering
  \includegraphics[width=1\linewidth]{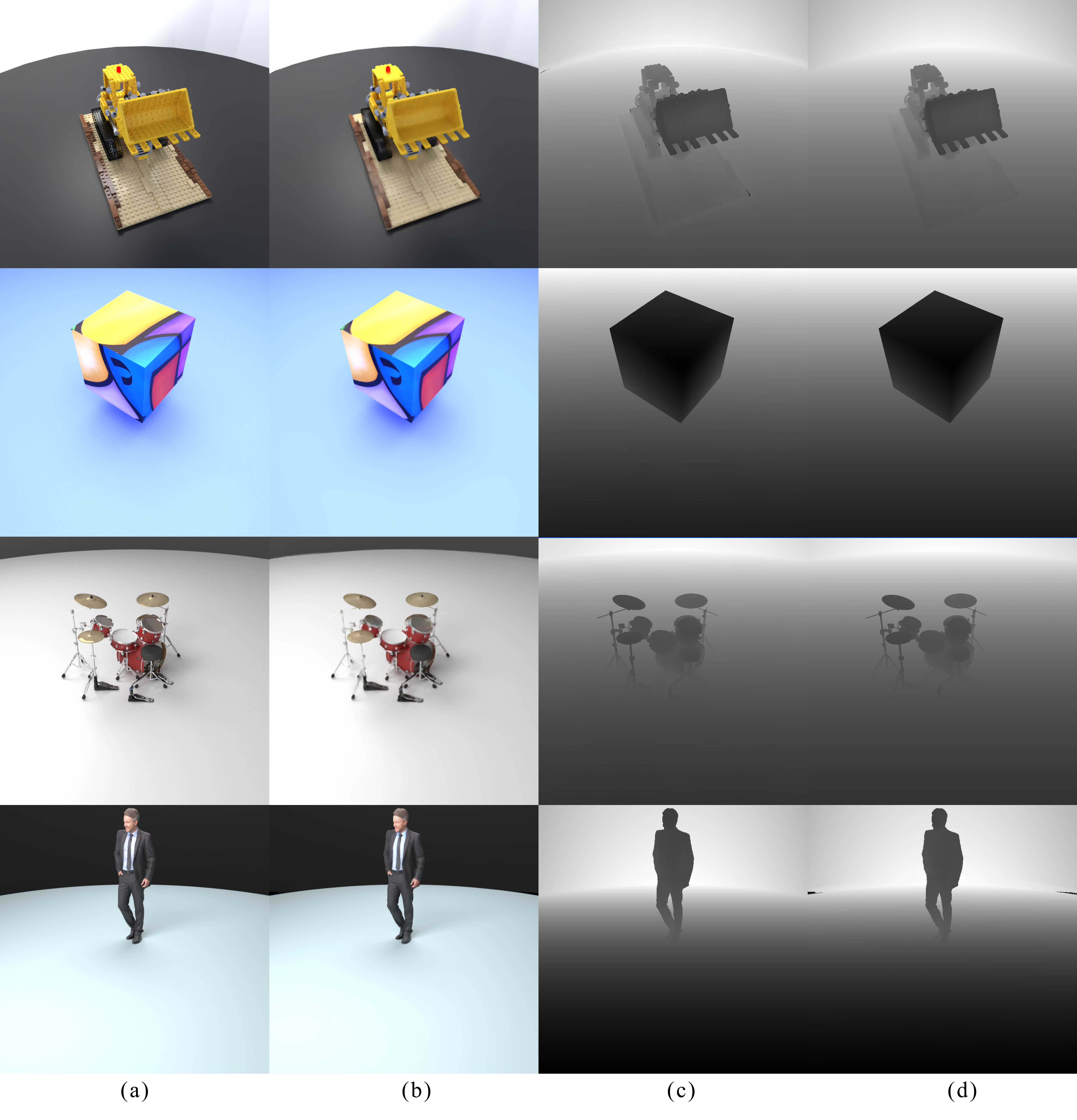}
  \caption{\label{fig:results}
          Qualitative results from simulated data. \textbf{(a)} Ground truth image;\textbf{(b)} Predicted image; \textbf{(c)} Ground truth depth; \textbf{(d)} Predicted depth.}
\end{figure}
\subsection{Stratified sampling} 
This approach is very similar to NeRF\cite{mildenhall2020nerf} sampling, the only difference is that the near and far bounds of the sampling are set using depth information. 
\subsection{Gaussian sampling}
Instead of placing the samples in a stratified manner, a Gaussian distribution is used to distribute sample locations around the surface. The mean of the distribution is the depth measurement, which ensures more samples are placed close to the surface.
\subsection{Adaptive sampling}
A multiview depth error map $\{e^i\}_{i=1}^N$ is generated using all depth maps of the training set (see the poster for the definition). The standard deviation of the Gaussian distribution is computed from $\{e^i\}_{i=1}^N$. It ensures that the spread of the samples are greater when there is more uncertainty in the depth.

\section{Preliminary Results}
\begin{table}[htb]
	\centering
	\begin{tabular}{|p{1cm}|p{1cm}|p{1cm}|p{1.3cm}|p{1.1cm}|}
	\hline
      & \multicolumn{4}{|c|}{Metrics} \\
    \hline
    dataset & PSNR$\uparrow$ & SSIM$\uparrow$ & Abs Rel$\downarrow$ & LPIPS$\downarrow$ \\ 
	\hline
	Lego & 27.4 & 0.933 & 0.012  & 0.0009 \\
	\hline
	Cube & 37.76 & 0.95 & 0.005 & 0.0001 \\
	\hline
	Drums  & 29.66 & 0.91 & 0.004 & 0.0008 \\
	\hline
	Human & 38.83 & 0.98 & 0.003  & 0.00006 \\
	\hline 
	\end{tabular}
	\caption{\label{table:datasets}The results of proposed method tested on 4 different simulated datasets. Underlying geometry is evaluated by Absolute relative distance(AbsRel). Photometric quality evaluated by PSNR(peak signal to noise ratio), SSIM(structural similarity index), and LPIPS(Learned Perceptual Image Patch Similarity).}
\end{table}
\begin{table}[htb]
	\centering
	\begin{tabular}{|p{1cm}|p{1cm}|p{1cm}|p{1cm}|p{1cm}|p{1cm}|}
	\hline
      & \multicolumn{5}{|c|}{Metrics} \\
    \hline
    Strategy & PSNR$\uparrow$ & SSIM$\uparrow$ & AbsRel$\downarrow$ & LPIPS$\downarrow$ & Time$\downarrow$ \\ 
	\hline
	Stratified & 21.81 & 0.891 & \textbf{0.003}  & 0.002 & 30m\\
	\hline
	Gaussian & \textbf{24.17} & \textbf{0.912} & 0.017 & 0.002 & \textbf{22m}\\
	\hline
	Adaptive & 23.40 & 0.910 & 0.018 & 0.002 & \textbf{22m}\\
	\hline
	NeRF & 22.3 & 0.84 & 0.215 & 0.002 & 1h 42m\\
	\hline
	\end{tabular}
	\caption{\label{table:sampling}The proposed local sampling strategies compared with baseline NeRF. The dataset contains 8 training images. Experiments were performed with 16 sample points.}
\end{table}
Qualitative results are shown in Table \ref{table:datasets} and \ref{table:sampling} . The Figure \ref{fig:results} shows qualitative results of the proposed local sampling. The best method according to the preliminary result is Gaussian sampling.
\section{Conclusions}
A preliminary study has been presented that shows that depth images can be used to perform local sampling and that fewer samples can reduce training time without compromising quality. The results suggest that surface information about the scenes can provide additional supervision to achieve better underlying geometry and photometry from fewer input views. 
\section{Acknowledgements}
The project is funded by Horizon 2020 COFUND BoostUrCAreer under the Marie Sklodowska-Curie grant agreement no. 847581 supported by the EU and the Region Sud-Provence-Alpes-Cote d'Azur and IDEX  ${UCA}^{{JEDI}}$. This work was granted access to the HPC resources of IDRIS under the allocation 2021-AD011012578 made by GENCI.

\bibliographystyle{unsrtnat}
\bibliography{references}  






\end{document}